\documentclass[10pt,twocolumn,letterpaper]{article}

\usepackage{cvpr}              

\usepackage[dvipsnames]{xcolor}
\usepackage{times}
\usepackage{epsfig}
\usepackage{graphicx}
\usepackage{amsmath}
\usepackage{amssymb}
\usepackage{booktabs}
\usepackage{multirow}
\usepackage{bm}
\usepackage{adjustbox}
\usepackage{pifont}
\usepackage{diagbox}
\usepackage{xcolor}
\usepackage{amsmath,mathtools}
\usepackage{algorithm}
\usepackage{cite}
\usepackage{color}
\usepackage{cuted}
\usepackage{etoolbox}
\usepackage {colortbl,array,xcolor}
\usepackage{datatool}
\usepackage{pgfplotstable}
\usepackage[hang,flushmargin]{footmisc}
\AtBeginDocument{
  \abovedisplayskip     =0.5\abovedisplayskip
  \abovedisplayshortskip=0.5\abovedisplayshortskip
  \belowdisplayskip     =0.5\belowdisplayskip
  \belowdisplayshortskip=0.5\belowdisplayshortskip
  }

\usepackage{tikz}
\definecolor{cvprblue}{rgb}{0.21,0.49,0.74}
\usepackage[pagebackref,breaklinks,colorlinks,citecolor=cvprblue]{hyperref}

\title{Face2Diffusion for Fast and Editable Face Personalization}

\author{\stepcounter{footnote}Kaede Shiohara
$\quad$ Toshihiko Yamasaki \\ 
The University of Tokyo\\ 
{\tt\small \{shiohara, yamasaki\}@cvm.t.u-tokyo.ac.jp}
}

\begin{document}

\maketitle

\begin{abstract}
Face personalization aims to insert specific faces, taken from images, into pretrained text-to-image diffusion models.
However, it is still challenging for previous methods to preserve both the identity similarity and editability due to overfitting to training samples.
In this paper, we propose Face2Diffusion (F2D) for high-editability face personalization.
The core idea behind F2D is that removing identity-irrelevant information from the training pipeline prevents the overfitting problem and improves editability of encoded faces.
F2D consists of the following three novel components:
1)~Multi-scale identity encoder provides well-disentangled identity features while keeping the benefits of multi-scale information, which improves the diversity of camera poses.
2)~Expression guidance disentangles face expressions from identities and improves the controllability of face expressions.
3)~Class-guided denoising regularization encourages models to learn how faces should be denoised, which boosts the text-alignment of backgrounds. 
Extensive experiments on the FaceForensics++ dataset and diverse prompts demonstrate our method greatly improves the trade-off between the identity- and text-fidelity compared to previous state-of-the-art methods.
Code is available at \url{https://github.com/mapooon/Face2Diffusion}.

\end{abstract}    
\begin{figure}[t]
    \centering
    \includegraphics[width=1.0\linewidth]{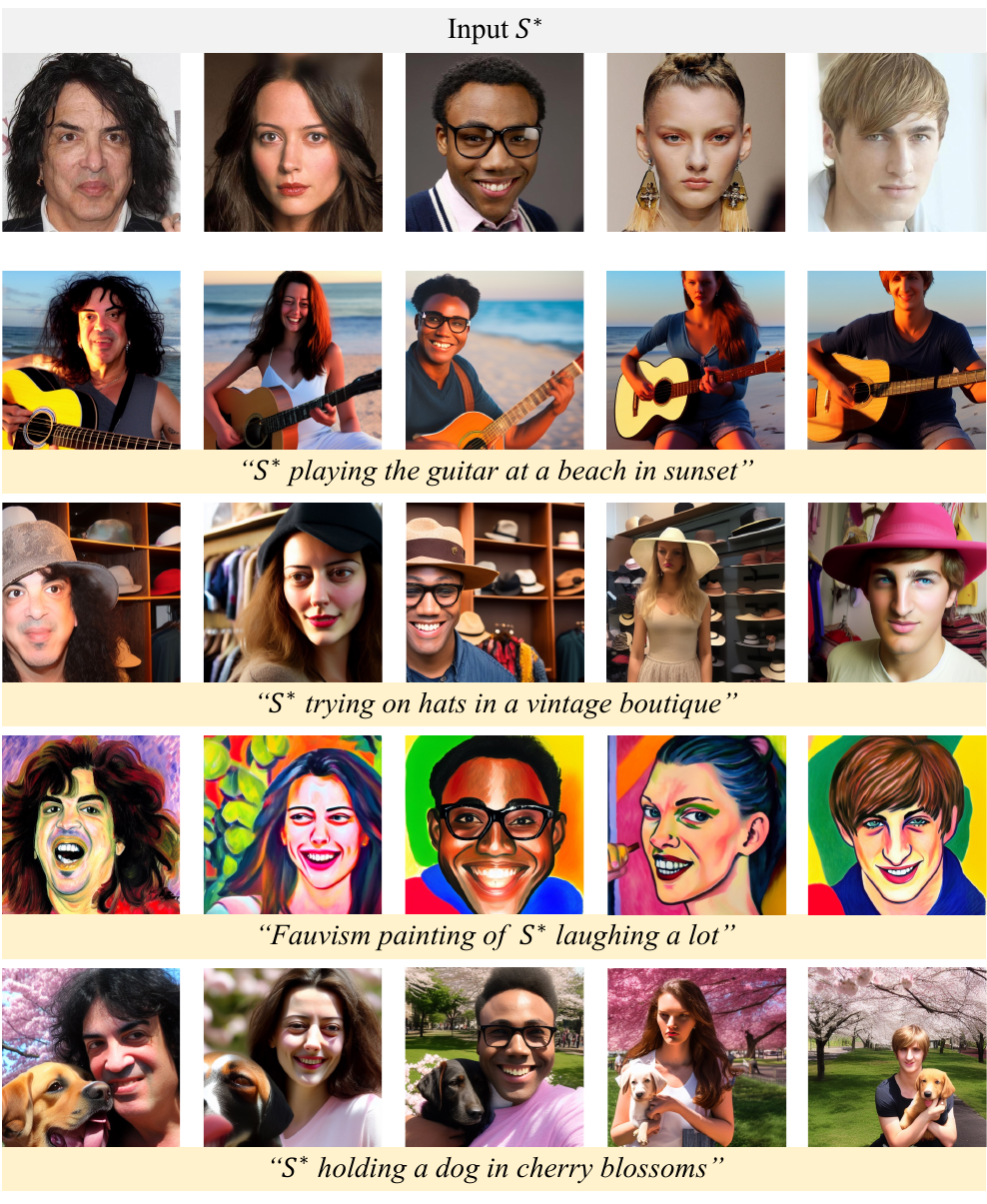}
    \vspace*{-0.5cm}
    \caption{\textbf{Our Results.} Face2Diffusion satisfies challenging text prompts that include multiple conditions while preserving input face identities without individual test-time tuning. 
    }
    \label{fig:teaser}
\end{figure}

\vspace{-3mm}
\section{Introduction}
\label{sec:intro}

Text-to-image (T2I) diffusion models trained on web-scale data such as GLIDE~\cite{glide}, DALL-E2~\cite{dalle2}, Imagen~\cite{imagen}, and StableDiffusion~\cite{latentdiffusion} have shown the impressive image generation ability aligned with a wide range of textual conditions, outperforming previous generative models (\eg,~\cite{stylegan,vqvae,biggan}).
Therefore, the next challenge of the community is to explore how to insert specific concepts (\eg, someone's face) from images that T2I models do not know, leading to the pioneering personalization methods such as TextualInversion~\cite{textualinversion} and DreamBooth~\cite{dreambooth}. 
However, they require heavy fine-tuning that takes several tens of minutes per concept.
To reduce the training costs and improve the editability, some studies~\cite{customdiffusion,perfusion} propose efficient training strategies that optimize mainly cross-attention layers in denoising UNet. 
Others use pretrained image encoders (\eg, CLIP~\cite{clip}) to represent new concepts as textual embeddings~\cite{disenbooth,e4t,celebbasis}. 
Most recently, some studies~\cite{fastcomposer,elite,dreamidentity} achieve tuning-free personalization via large-scale pretraining.
For example, ELITE~\cite{elite} proposes a two-stage training strategy that optimizes its mapping networks and cross-attention layers
on the OpenImages dataset~\cite{openimages}.

In particular, personalization for faces, which we call ``face personalization'' in this paper, draws attention from the community because of its potential applications, \eg, content creation.
CelebBasis~\cite{celebbasis} represents face embeddings on a basis constructed by celebrity names known to T2I models.
FastComposer~\cite{fastcomposer} proposes localizing cross-attention to solve the identity blending problem where T2I models mix-up multiple people in one image.
DreamIdentity~\cite{dreamidentity} improves the identity similarity of generated images by utilizing multi-scale features from a pretrained face recognition model~\cite{arcface}.
However, despite the great efforts in personalization of T2I models, we found that it is still challenging to generate well-aligned images with a wide range of text prompts while preserving the subjects' identities because of overfitting to training images.

In this paper, we propose Face2Diffusion (F2D) for more editable face personalization.
The core idea behind our F2D is that properly removing identity-irrelevant information from the training pipeline helps the model learn editable face personalization.
We analyze the overfitting problem on a previous method~\cite{customdiffusion} and classify the typical overfitting into three types: camera poses, face expressions, and backgrounds, and present the following solutions for the problem:
1) Multi-scale identity (MSID) encoder is presented to provide purer identity features disentangled from camera pose information while keeping the benefits of multi-scale feature extraction.
2) Expression guidance disentangles face expressions from identity features, making it possible to control face expressions by text prompts and reference images.
These techniques enable F2D to disentangle backgrounds, camera poses, and face expressions, performing high-fidelity face personalization as shown in Fig.~\ref{fig:teaser}. 
3) Class-guided denoising regularization (CGDR) 
constrains backgrounds of injected faces to be denoised in the same manner as its super-class word (\ie, \textit{``a person''} for face personalization) to improve the text-fidelity on backgrounds.

\begin{figure}[t]
    \centering
    \includegraphics[width=1.0\linewidth]{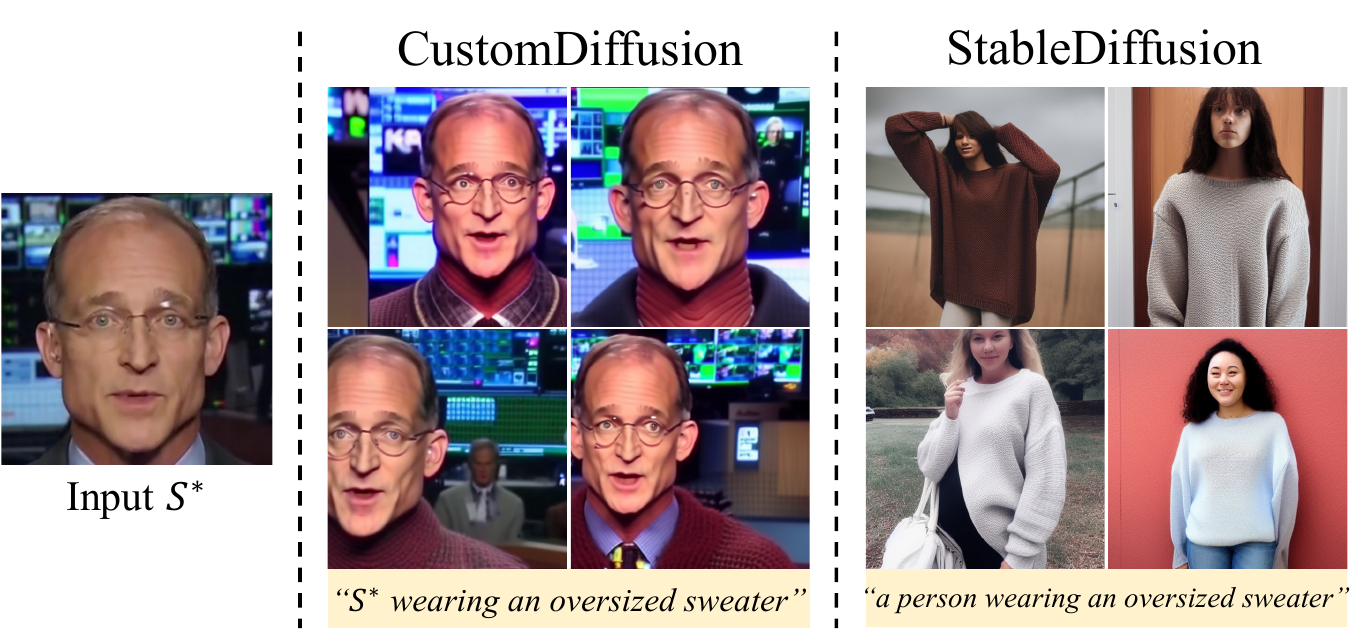}
    \vspace{-4.5mm}
    \caption{\textbf{Typical overfitting to input data.} The original StableDiffusion~\cite{latentdiffusion} is capable of generating text-aligned images with plausible backgrounds, camera poses, and diverse face expressions. Nevertheless, a previous method~\cite{customdiffusion} fails in disentangling these identity-irrelevant information from the input sample.}
    \label{fig:overfitting}
\end{figure}

We compare our F2D with the nine previous state-of-the-art methods~\cite{textualinversion,dreambooth,customdiffusion,perfusion,e4t,celebbasis,fastcomposer,elite,dreamidentity} on 100 faces from the FaceForensics++~\cite{ffpp} dataset with 40 diverse text prompts we collected.
Our method ranks in the top-3 in five of the six metrics 
in terms of the identity-fidelity~\cite{adaface,sphereface,facenet} and text-fidelity~\cite{clipscore,nada,siglip}, and outperforms previous methods in the harmonic and geometric means of the six metrics, reflecting the superiority of F2D in the total quality of face personalization.

\section{Related Work}
\label{sec:related_work}
\subsection{Text-to-Image Diffusion Models.}
Text-to-image (T2I) diffusion models~\cite{latentdiffusion,glide,imagen,dalle2} are a kind of generative models based on diffusion models~\cite{sohl2015deep,ddpm,ddim} that generate images by gradual denoising steps in the image space~\cite{glide,imagen,dalle2} or latent one~\cite{latentdiffusion}.
Specifically, T2I models generate images from encoded texts by a large language model~\cite{t5} or vision-language model~\cite{clip}.
ADM~\cite{adm} guides denoised images to be  desired classes using image classifiers. 
Classifier-free guidance~\cite{cfdg} achieves to guide diffusion models without any classifiers, overcoming the limitation of guided diffusion that requires noise-robust image classifiers.

\subsection{Personalization of T2I models}
The high-fidelity generation ability of T2I models drives the community to explore personalization methods, or how to insert unseen concept into T2I models. 
Here, we review existing works by classifying them into three types:

\noindent\textbf{Direct optimization methods.}
Early attempts optimize a learnable word embedding for a new concept~\cite{textualinversion} or finetune the whole T2I model~\cite{dreambooth} by learning to reconstruct input images from a new concept word embedding. 
CustomDiffusion~\cite{customdiffusion} identifies that cross-attention layers of the denoising UNet are more influential to invert new concepts and proposes a lightweight fine-tuning method. 
Perfusion~\cite{perfusion} replaces the key value of a new concept embedding with that of its super-category's one in the cross-attention layers to mitigate overfitting to training samples.

\noindent\textbf{Encoder-based optimization methods.}
To improve editability and training efficiency, some studies~\cite{e4t,disenbooth,celebbasis} propose to utilize pretrained image encoders such as CLIP~\cite{clip} to represent new concepts as textual embeddings.
E4T~\cite{e4t} proposes a two-stage personalization method that includes pretraining of a word embedder and weight offsets of attention layers on domain-oriented datasets~\cite{stylegan,progan,lsun} and few-step (5-15) fine-tuning per concept.
DisenBooth~\cite{disenbooth} learns to disentangle a new concept from irrelevant information by reconstructing an concept image from another image.

\noindent\textbf{Optimization-free methods.}
Motivated by the impressive results of the encoder-based methods, recent studies~\cite{instantbooth,elite,fastcomposer,dreamidentity,subjectdiffusion} focus on feed-forward inversion without fine-tuning per concept.
ELITE~\cite{elite} introduces a two-stage pretraining framework that optimizes global and local mapping networks and cross-attention layers of its T2I model.
FastComposer~\cite{fastcomposer} replaces the inserted embeddings of input text prompts with their super-class embeddings in early denoising steps during inference to improve the text-fidelity.
DreamIdentity~\cite{dreamidentity} leverages generated images by its T2I model to train itself on various image-text pairs without manual collection.

\begin{figure*}[t]
    \centering
    \includegraphics[width=1.0\linewidth]{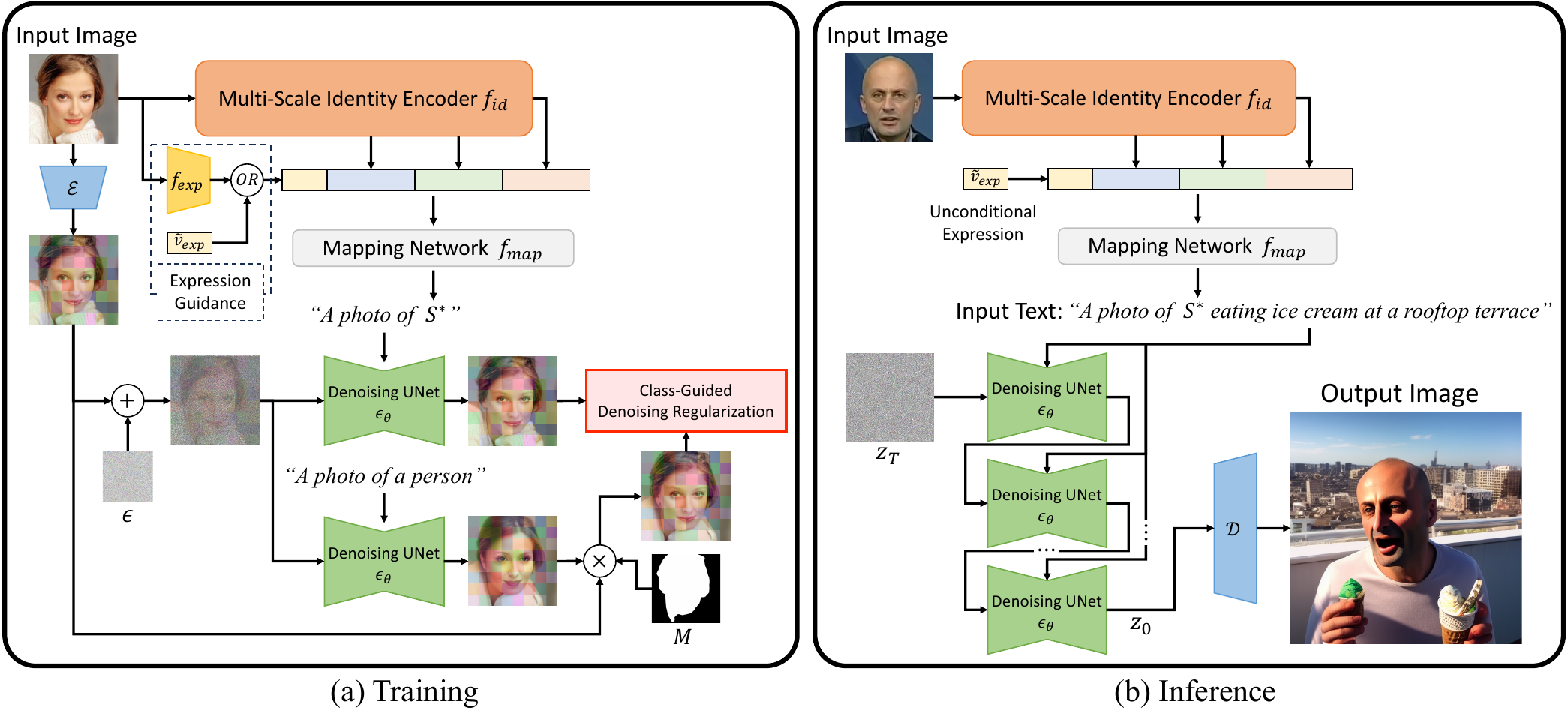}
    \caption{\textbf{Overview of Face2Diffusion.} (a) During training, we input a face image into our novel multi-scale identity encoder~$f_{id}$ and an off-the-shell 3D face reconstruction model~$f_{exp}$ to extract identity and expression features, respectively. The concatenated feature is projected into the text space as a word embedding $S^*$ by a mapping network $f_{map}$. The input image is also encoded by VAE's encoder~$\mathcal{E}$ and then a Gaussian noise $\epsilon$ is added to it. 
    We constrain the denoised latent feature map to be the original one in the foreground and to be a class-guided denoised result in the background. 
    (b) During inference, the expression feature is replaced with an unconditional vector $\tilde{v}_{exp}$ to diversify face expressions of generated images. After injecting the face embedding $S^*$ into an input text, the original denoising loop of StableDiffusion is performed to generate an image conditioned by the input face identity and text. 
    }
    \label{fig:overview}
\end{figure*}

\section{Face2Diffusion}
Our goal is to represent input faces as face emeddings $S^*$ in the text space of CLIP~\cite{clip} to generate target subjects conditioned by text prompts on StableDiffusion~\cite{latentdiffusion}. 
The important observation behind our work is that previous methods suffer from three types of overfitting: backgrounds, camera poses, and face expressions.
Fig.~\ref{fig:overfitting} shows a failure case of a previous method~\cite{customdiffusion}. It can be observed that the method tends to generate similar backgrounds, camera poses, and face expressions due to overfitting to the input sample despite the capability of the original StableDiffusion generating suitable and diverse scenes.
Motivated by this overfitting problem, we propose Face2Diffusion (F2D) for more editable face personalization.
We visualize the overview of F2D in Fig.~\ref{fig:overview}.
F2D consists of three important components to solve the overfitting promblem: 
In Sec.~\ref{sec:msid}, we introduce multi-scale identity (MSID) encoder to disentangle camera poses from face embeddings by removing identity-irrelevant information from a face recognition model~\cite{arcface}.
In Sec.~\ref{sec:eg}, we present expression guidance that disentangles face expressions from face embeddings to diversify face expressions aligned with texts.
In Sec.~\ref{sec:cgdr}, we present class-guided denoising regularization (CGDR) that forces the denoising manner of face embeddings to follow that of its super-class (\ie, \textit{``a person''}) in the background.

\subsection{Preliminary}
\noindent\textbf{StableDiffusion.}
We adopt StableDiffusion~\cite{latentdiffusion} (SD) as our base T2I model.
SD first learns perceptional image compression by VAE~\cite{vae}. 
Then, SD learns conditional image generation via latent diffusion.
Given a timestep $t$, noisy latent feature $z_t$, and text feature $\tau(p)$ where $\tau$ and $p$ are the CLIP text encoder~\cite{clip} and a text prompt, respectively, a denoising UNet~\cite{unet} $\epsilon_{\theta}$ predicts the added noise $\epsilon$. The training objective is formulated as:
\begin{equation}
\label{eq:loss_rec}
\mathcal{L}_{ldm} = \lVert \epsilon - \epsilon_{\theta}(z_t,t,\tau(p)) \rVert_2^2.
\end{equation}
Leveraging large-scale image-text pair datasets (\eg, LAION-5B~\cite{laion}), SD learns semantic relationships between images and texts.

\noindent\textbf{Encoder-based face personalization.}
Previous works~\cite{celebbasis,dreamidentity,fastcomposer,e4t} show that pretrained image encoders are usefull to represent face identities as textual embeddings $S^*$. 
For a face encoder $f_{id}$ and a mapping network $f_{map}$,
$S^*$ for an input image $x$ is computed as follows:
\begin{equation}
\label{eq:encoding}
S^* = f_{map}(f_{id}(x)).
\end{equation}
Then, $S^*$ is used as a tokenized word to generate identity-injected images as in Fig.~\ref{fig:overview} (b). $f_{map}$ is optimized using the reconstruction loss Eq.~\ref{eq:loss_rec} per subject~\cite{celebbasis} or face datasets for instant encoding without test-time tuning~\cite{fastcomposer,dreamidentity}.

\subsection{Multi-Scale Identity Encoder}
\label{sec:msid}
We start by considering how to encode faces in the text space.
Previous works~\cite{celebbasis,dreamidentity} reveal that face recognition models~\cite{arcface,cosface} provide beneficial identity representations for face personalization. 
Specifically, DreamIdentity~\cite{dreamidentity} proposes to extract multi-scale features from shallow layers to deep ones to improve the identity similarity.

\begin{figure}[t]
    \centering
    \begin{minipage}[b]{0.49\linewidth}
    \centering
    \includegraphics[width=1.0\linewidth]{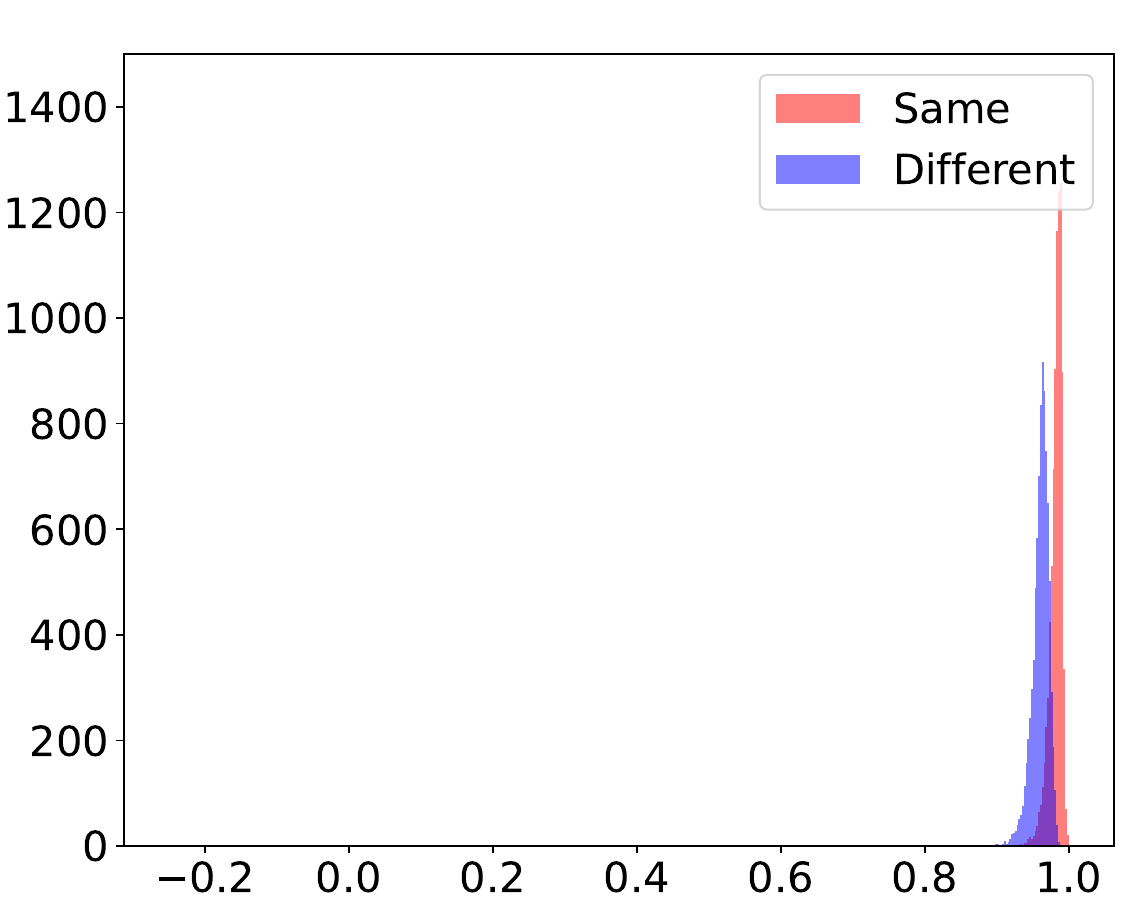}
    \subcaption{3rd layer of ArcFace}
    \label{arcface_3}
    \end{minipage}
    \begin{minipage}[b]{0.49\linewidth}
    \centering
    \includegraphics[width=1.0\linewidth]{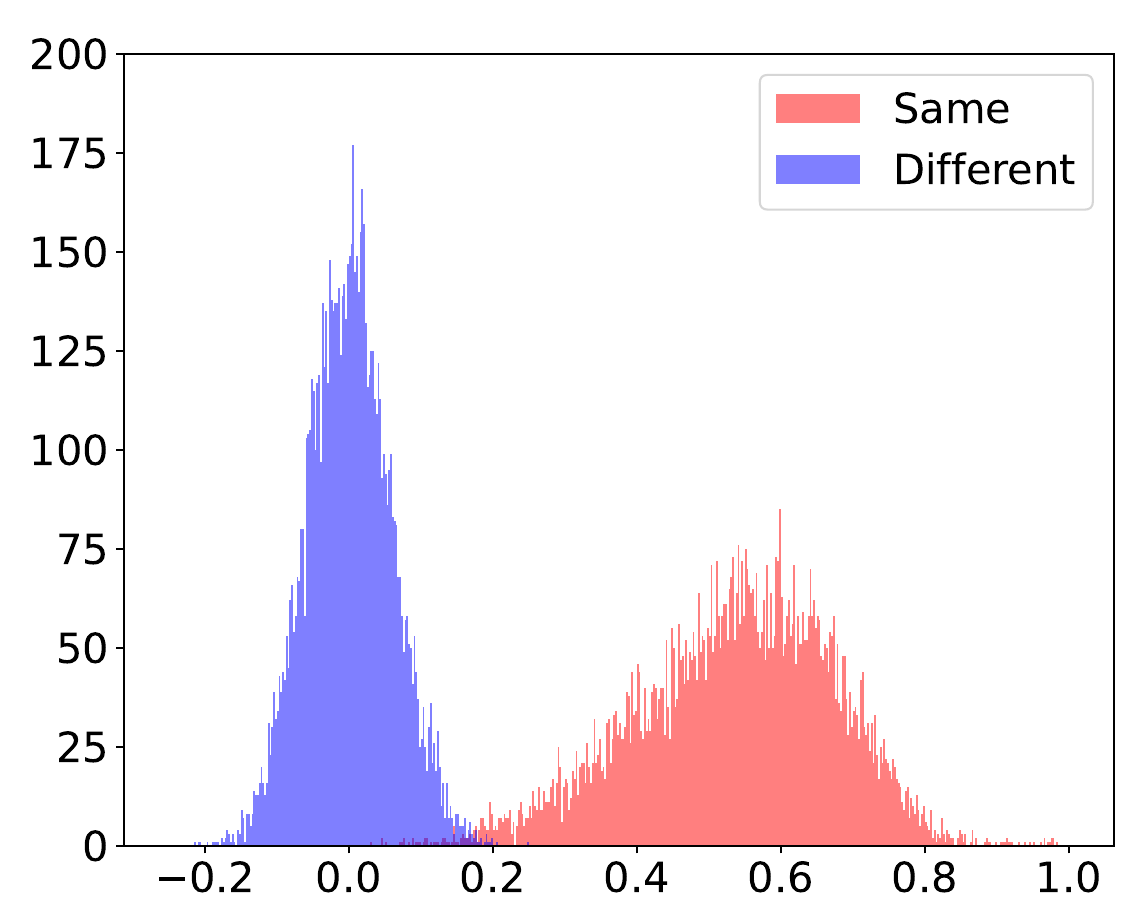}
    \subcaption{12th layer of ArcFace}
    \label{arcface_12}
    \end{minipage}\\
    \begin{minipage}[b]{0.49\linewidth}
    \centering
    \includegraphics[width=1.0\linewidth]{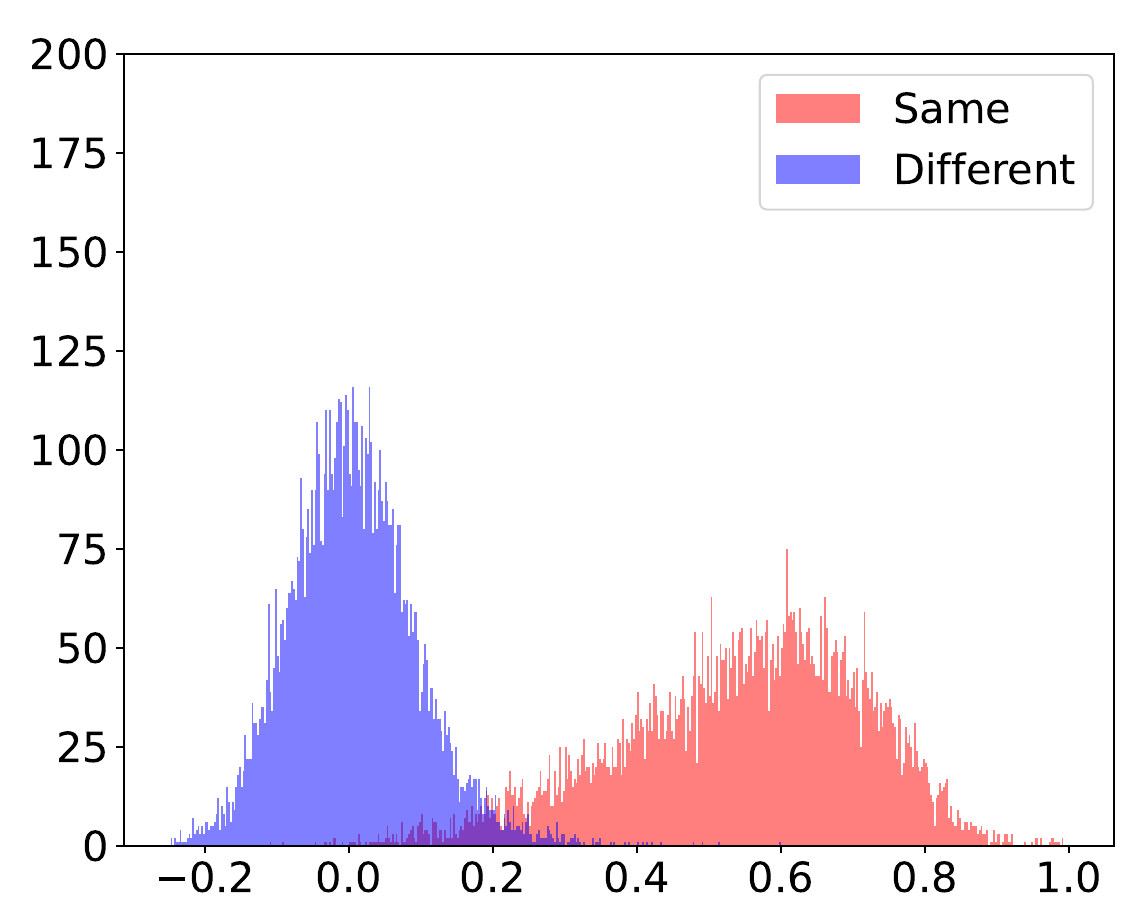}
    \subcaption{3rd layer of MSID encoder}
    \label{msid_3}
    \end{minipage}
    \begin{minipage}[b]{0.49\linewidth}
    \centering
    \includegraphics[width=1.0\linewidth]{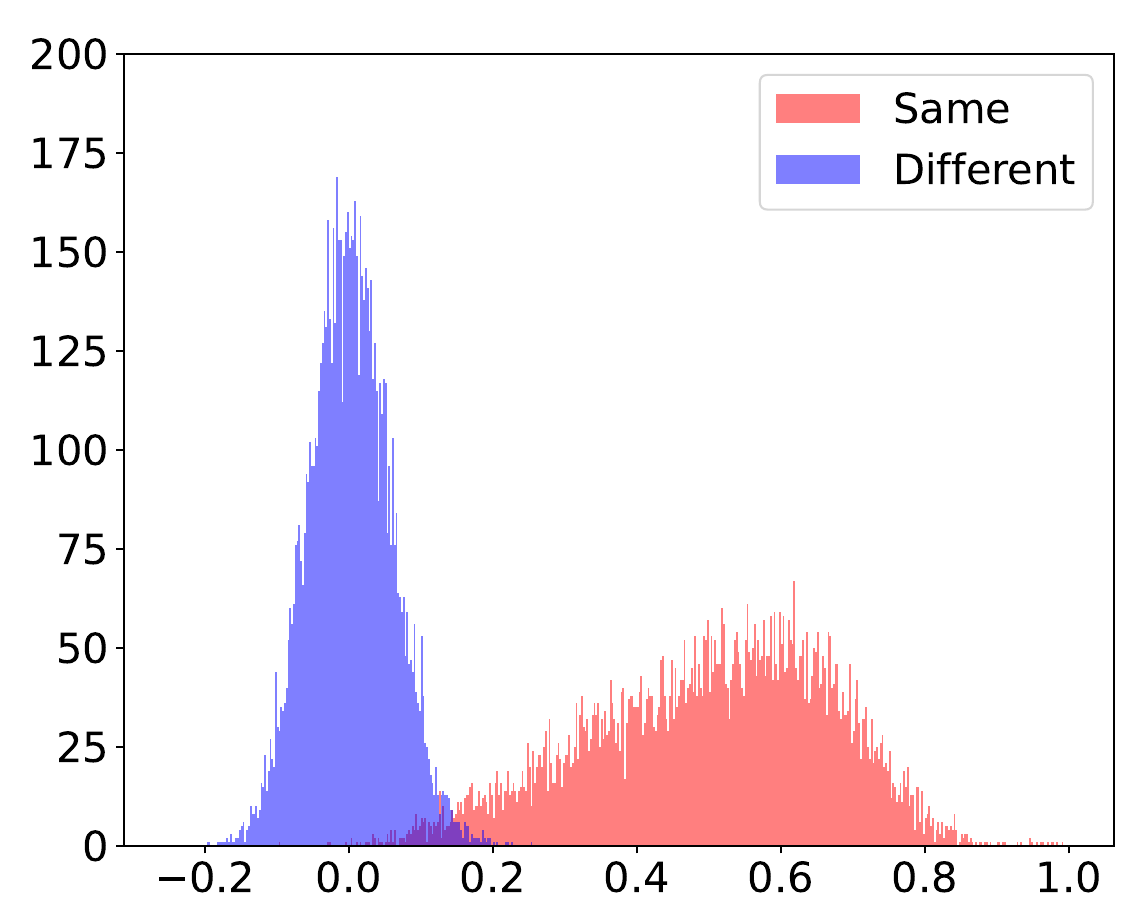}
    \subcaption{12th layer of MSID encoder}
    \label{msid_12}
    \end{minipage}
    \caption{\textbf{Identity similarity distributions of ArcFace and MSID encoder.} Shallow features of ArcFace (a) are un-aligned and low discriminability although deep features (b) are well aligned. Our MSID encoder succeeds in removing identity-irrelevant information both from the shallow layer (c) and deep one (d) to prevent overfitting.}
    \label{fig:mlvit_vs_vit}
\end{figure}

However, such a straight-forward multi-scale method suffers from overfitting to input samples. 
This is because features from shallow layers include a lot of low level information (\eg, camera pose, expression, and background) that is irrelevant to identity~\cite{elite}. 
This accidentally makes models dependent on shallow features to minimize the reconstruction loss Eq.~\ref{eq:loss_rec}, which results in a low editability.
Figs.~\ref{fig:mlvit_vs_vit}(a) and (b) visualize identity similarity distributions of ViT~\cite{vit}-based ArcFace~\cite{arcface} on a shallow (3rd) layer and the deepest (12th) one, respectively.
The horizontal and vertical axes represent the identity similarity and frequency, respectively.
We uniformly sample an \{Anchor, Positive (Same), Negative (Different)\} triplet for each identity from the VGGFace2~\cite{vggface2} dataset.
It can be observed that the shallow features are un-aligned and insufficient to discriminate identities; AUC on shallow features is only 93.99\% although that on the deepest is 99.97\%.
This result implies that shallow features contain specific information of input images, which causes overfitting in face personalization.

To solve this problem, we present a multi-scale identity (MSID) encoder for face personalization.
In the pre-training of a face recognition model ArcFace~\cite{arcface}, we encourage not only the deepest layer but also shallower ones to discriminate identities. 
We denote the concatenated multi-scale identity vector as $v_{id}=\lbrack v_{id}^{1},v_{id}^{2},...,v_{id}^{D}\rbrack$ where $v_{id}^{i}$ is the $i$-th level feature.
Given an image and its class label $y$, we formulate our multi-scale loss $\mathcal{L}_{m}$ as the original ArcFace loss~\cite{arcface}:
\begin{equation}
\label{loss_pretraining}
\mathcal{L}_{m} = - \log \frac{e^{s\cos(W^T_{y}v_{id} + m)}}{e^{s\cos(W^T_{y}v_{id} + m)}+\sum_{k=1, k\ne y}^{K}e^{s\cos W^T_{k}v_{id}}},
\end{equation}
where $W_{k}$ is the $k$-th column of the weight matrix $W$.  
$m$ and $s$ are the margin and scale of ArcFace~\cite{arcface}, respectively.
As shown in Figs.~\ref{fig:mlvit_vs_vit}(c) and (d), the multi-scale features by our encoder are well-aligned and discriminative both on shallow and deep layers; AUC on shallow features is improved from 93.99\% to 99.46\%.
The disentangled multi-scale identity features prevent overfitting while providing strong identity information for face personalization.
Differently from a previous work in image classification tasks~\cite{arcfusion} that targets only two deepest layers, we train a wide range of levels because we focus on the disentanglement of multi-scale features from shallow to deep layers for face personalization, rather than the discriminability of the deepest feature for classification.

\subsection{Expression Guidance}
\label{sec:eg}
Although our MSID encoder greatly improves the text-fidelity while preserving the identity similarity, we found that it is still difficult to control face expressions due to overfitting to input samples.
To tackle this problem, we propose expression guidance to disentangle face expressions from identities.
We utilize a 3D face reconstruction model~\cite{deep3dfacerecon}, denoted as $f_{exp}$, to extract expression features.

For an input face image $x$, we simply inject expression information by concatenating $v_{exp}=f_{exp}(x)$ and $v_{id}=f_{id}(x)$ before the mapping network $f_{map}$.
The expression-guided feature vector $S^*$ is computed as:
\begin{equation}
  S^* = 
  \begin{cases}
    f_{map}([v_{id}, v_{exp}]) & (p=0.8), \\
    f_{map}([v_{id}, \tilde{v}_{exp}]) & (p=0.2),
  \end{cases}
\end{equation}
where we drop $v_{exp}$ with a probability of 0.2 and input an alternative learnable vector $\tilde{v}_{exp}$ for unconditional generation. 
After training, we generate target subjects aligned with diverse expression-related prompts using the learned unconditional vector $\tilde{v}_{exp}$.
Accessorily, F2D can perform conditional generation with expression references $v_{exp}$. Please see the supplementary material for more details.

\begin{figure*}[t]
    \centering
    \includegraphics[width=1.0\linewidth]{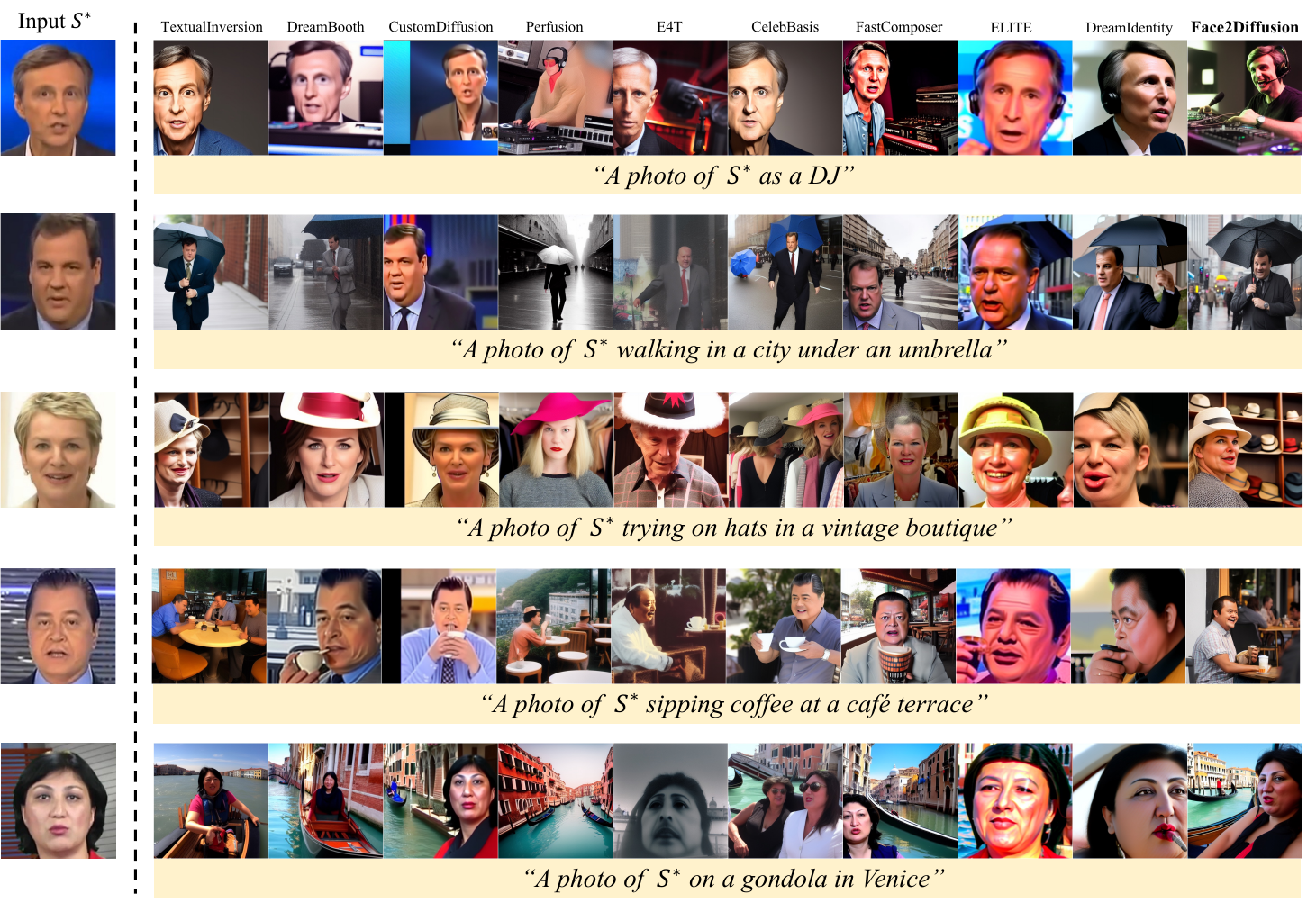}
    \vspace*{-0.6cm}
    \caption{\textbf{Qualitative comparison with previous methods.} Our method generates authentic images aligned with input texts and identities in challenging scenes whereas previous methods compromise an either. Enlarged images are included in the supplementary materials.}
    \label{fig:comparison}
\end{figure*}

\subsection{Class-Guided Denoising Regularization}
\label{sec:cgdr}
Overfitting to training samples sometimes prevent generated images from aligned with text conditions in the background.
Delayed subject conditioning (DSC)~\cite{fastcomposer} tackles this problem by replacing the identifier $S^*$ with its super-class word (\eg, \textit{``a person''}) in early denoising steps and switching the super-class word with $S^*$ from the middle of the denoising loop during inference. 
However, because some of the face identities emerge in early denoising steps, DSC degrades the identity similarity of generated images to input faces.
To disentangle background information from face embeddings without identity degradation, we propose class-guided denoising regularization (CGDR).
The important observation behind CGDR is that the class word \textit{``a person''} is well disentangled from specific backgrounds as various ones are generated in Fig.~\ref{fig:overfitting}.
CGDR forces the denoising manner of face embeddings to follow that of the class prompt in the background.
First, we input an identity-injected prompt $p=\textit{``A photo of } S^*\textit{''}$ and class prompt $p_c=\textit{``A photo of a person''}$ into our model to predict the noise for the same noisy latent $z_t$:
\begin{equation}
 \hat{\epsilon}=\epsilon_{\theta}(z_t,t,\tau(p)),
\end{equation}
\begin{equation}
 \hat{\epsilon_c}=\epsilon_{\theta}(z_t,t,\tau(p_c)).
\end{equation}
Then, we constrain the original predicted noise $\hat{\epsilon}$ to be $\hat{\epsilon}_c$ in the background and to be the actually added noise $\epsilon$ in the foreground as follows:
\begin{equation}
\label{eq:loss_cgdr}
 \mathcal{L}=\lVert \hat{\epsilon} - \{\epsilon \odot M + \hat{\epsilon}_c \odot (1-M)\} \rVert_2^2,
\end{equation}
where $M$ is a segmentation mask by a face-parsing model~\cite{faceparsing} to divide the identity region from others and $\odot$ represents pixel-wise multiplication.
By this single objective function, F2D learns to encode face identities, strongly mitigating overfitting to the backgrounds of input images.

Notably, our training strategy keeps the weights of T2I models original because we optimize only $f_{map}$ during training of F2D. 
Therefore, our method fundamentally does not struggle with the language drift~\cite{drift} and catastrophic forgetting~\cite{catastrophic} encountered by some of previous methods (\eg,~\cite{dreambooth,disenbooth,instantbooth,fastcomposer}) that modify the original weights.

\begin{table*}[t]
    \centering
    \begin{adjustbox}{width=0.85\linewidth}
    \begin{tabular}{lcccccc|cc} \toprule
      \multirow{2}{*}{Method} & \multicolumn{3}{c}{Identity-Fidelity ($\uparrow$)} & \multicolumn{3}{c}{Text-Fidelity ($\uparrow$)} &\multicolumn{2}{c}{Identity$\times$Text ($\uparrow$)} \\
        \cmidrule(lr){2-4} \cmidrule(lr){5-7}\cmidrule(lr){8-9}
      & AdaFace &SphereFace & FaceNet & CLIP&dCLIP &SigLIP&hMean&gMean\\
      \midrule
        \multicolumn{4}{l}{\textit{Subject-driven}}\\
        TextualInversion~\cite{textualinversion} & 0.1654 & 0.2518 & 0.3197 & 0.1877 & 0.1075 & 0.1283 & 0.0320 & 0.0575\\
        DreamBooth~\cite{dreambooth} & 0.3482(2) & 0.4084 & 0.4774 & 0.2351 & 0.1480 & 0.3452(3) & 0.0689 & 0.1116\\
        CustomDiffusion~\cite{customdiffusion} & 0.4537(1) & 0.5624(1) & 0.6458(1) & 0.2023 & 0.1490 & 0.2182 & 0.1078 & 0.1700(3)\\
        Perfusion~\cite{perfusion} & 0.0925 & 0.1478 & 0.1887 & 0.2490 & 0.1686 & 0.3433 & 0.0342 & 0.0575\\
        E4T~\cite{e4t} & 0.0433 & 0.1190 & 0.1725 & 0.2510(2) & 0.1784 & 0.2671 & 0.0370 & 0.0589\\
        CelebBasis~\cite{celebbasis} & 0.1601 & 0.2724 & 0.3762 & 0.2563(1) & 0.1847(3) & 0.3624(2) & 0.1138(3) & 0.1542\\
        \midrule
        \multicolumn{4}{l}{\textit{Subject-agnostic}}\\
        FastComposer~\cite{fastcomposer} & 0.1736 & 0.3271 & 0.4725 & 0.2495(3) & 0.2149(1) & 0.3168 & 0.1655(2) & 0.2120(2)\\
        ELITE~\cite{elite} & 0.0924 & 0.1925 & 0.3232 & 0.1755 & 0.1250 & 0.0671 & 0.0308 & 0.0624\\
        DreamIdentity$^*$~\cite{dreamidentity} & 0.2847 & 0.4252(2) & 0.5523(2) & 0.1924 & 0.1463 & 0.1539 & 0.0849 & 0.1326\\
        \rowcolor[gray]{0.9}
        Face2Diffusion (Ours) & 0.3143(3) & 0.4215(3)& 0.5313(3)& 0.2486 & 0.2020(2) & 0.3856(1) & 0.1749(1) & 0.2252(1)\\
      \bottomrule
    \end{tabular}
    \end{adjustbox}
  \caption{\textbf{Comparison with previous methods.} The top-3 values are ranked in brackets. $^*$ denotes our own implementation.  CustomDiffusion~\cite{customdiffusion} suffers from overfitting to input faces while FastComposer~\cite{fastcomposer} cannot represent identities sufficiently, resulting in lower Identity$\times$Text scores. Our model ranks in the top-3 in five of the six metrics and achieves the best Identity$\times$Text scores.}
  \label{tb:ffpp}
\end{table*}

\section{Experiments}
\subsection{Implementation Detail}
\noindent\textbf{Multi-scale identity encoder.}
We pretrain ViT~\cite{vit} equipped with 12 transformer encoder layers on the MS1M~\cite{ms1m} dataset for 30 epochs.
The batch size and learning rate are set to $1024$ and $10^{-3}$, respectively.
We select a depth set of $\{3,6,9,12\}$ for multi-scale feature extraction.

\noindent\textbf{Face2Diffusion.}
We adopt the two-word embedding method~\cite{celebbasis,dreamidentity} where $S^*=[S_1^*, S_2^*]$ is predicted by two independent mapping networks $f_{map}^1$ and $f_{map}^2$.
Each mapping network $f_{map}^{i}$ consists of two-layer MLP where each layer has a linear layer, dropout~\cite{dropout}, and leakyReLU~\cite{lrelu}, followed by an additional linear layer that projects the features into the text space.
We train our model on FFHQ~\cite{stylegan} for 100K iterations. 
Note that only $f_{map}^1$ and $f_{map}^2$ are updated and other networks are frozen.
Only horizontal flip is used for data augmentation.
The batch size and learning rate are set to $32$ and $10^{-5}$, respectively.

\subsection{Setup}
\noindent\textbf{Evaluation data.}
We adopt the FaceForensics++~\cite{ffpp} dataset. We randomly select 100 videos and then extract a frame from each video. We align and crop images following the FFHQ~\cite{stylegan} dataset.
We also collect 40 human-centric prompts including diverse scenes such as job, activity, expression, and location. 
The prompt list is included in the supplementary material.
We generate all combinations of face images and prompts, \ie, 4,000 scenes.
Following previous works~\cite{dreambooth,disenbooth}, we generate four images for each \{image, prompt\} set for robust evaluation.
We report averaged scores over all samples.

\noindent\textbf{Compared methods.}
We compare our method with nine state-of-the-art personalization methods including six subject-driven methods, TextualInversion~\cite{textualinversion}, DreamBooth~\cite{dreambooth}, CustomDiffusion~\cite{customdiffusion}, Perfusion~\cite{perfusion}, E4T~\cite{e4t}, and CelebBasis~\cite{celebbasis}, and three subject-agnostic methods, FastComposer~\cite{fastcomposer}, ELITE~\cite{elite}, and DreamIdentity~\cite{dreamidentity}.
For all the methods including our Face2Diffusion, we use StableDiffusion-v1.4~\cite{latentdiffusion}, Euler ancestral discrete scheduler~\cite{kdiffusion} with 30 denoising steps, and classifier-free guidance~\cite{cfdg} with a scale parameter of $7.0$ for fair comparison.
Please refer to the supplementary material for more details.

\begin{table}[t]
    \centering
    \begin{adjustbox}{width=1.0\linewidth}
    \begin{tabular}{lcccc} \toprule
        &CustomDiffusion& CelebBasis & FastComposer & Ours\\ 
      \midrule
      FID ($\downarrow$)&86.18&\underline{69.87}&77.62&\textbf{69.33}\\
      \#Params ($\downarrow$)& $5.71 \times 10^7$ &$\textbf{1024}$& $8.88 \times 10^8$  &\underline{$1.20 \times 10^7$} \\ 
      Time ($\downarrow$)& 140 sec &220 sec& \underline{0.026 sec} &\textbf{0.006 sec} \\
      \bottomrule
    \end{tabular}
    \end{adjustbox}
  \caption{\textbf{Comparison on FID and the computation costs.}}
  \label{tb:param_fid}
\end{table}

\noindent\textbf{Metrics.}
We consider three evaluation metrics following:

\noindent 1) Identity-fidelity represents how a generated image is similar to an input image in terms of face identity.
We adopt AdaFace~\cite{adaface}, SphereFace~\cite{sphereface}, and FaceNet~\cite{facenet}, and compute the cosine similarity between extracted features from input images and generated images. 
We assign $0.0$ to images where no face is detected.

\noindent 2) Text-fidelity expresses how a generated image is aligned with a text prompt except the face identity.
We adopt CLIP score~\cite{clipscore}, directional CLIP (dCLIP) score~\cite{nada}, and SigLIP~\cite{siglip} to compute image-text similarity.
We use a reference prompt ``\textit{A photo of a person}'' and a reference image generated by ``\textit{A photo of $S^*$}'' for dCLIP.

\noindent 3) Because the goal of face personalization is to satisfy ``injecting a face identity'' and ``being aligned with a text prompt'' simultaneously, it is important to evaluate these performance simultaneously on each image.
To meet this demand, we introduce Identity$\times$Text score that reflects the total quality of face personalization. 
We compute the harmonic mean (hMean) and geometric one (gMean) of the six metrics above.

\begin{figure*}[t]
  \centering
  \begin{minipage}{.33\textwidth}
    \centering
    \includegraphics[height=5.75cm]{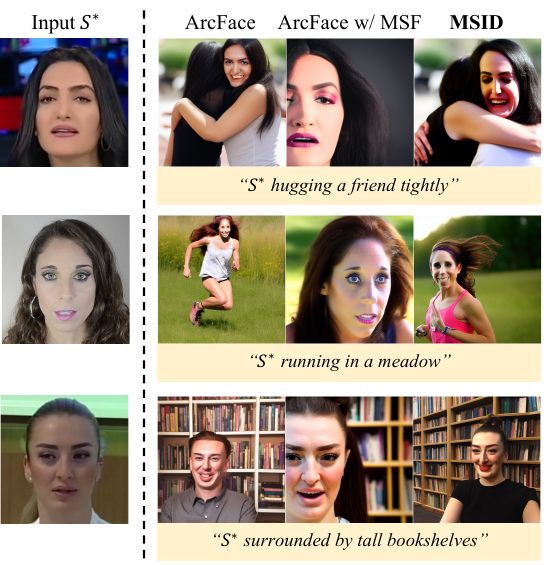} 
    \subcaption{Effect of MSID encoder.}
    \label{fig:ablation_encoder}
  \end{minipage}
  \begin{minipage}{.23\textwidth}
    \centering
    \includegraphics[height=5.75cm]{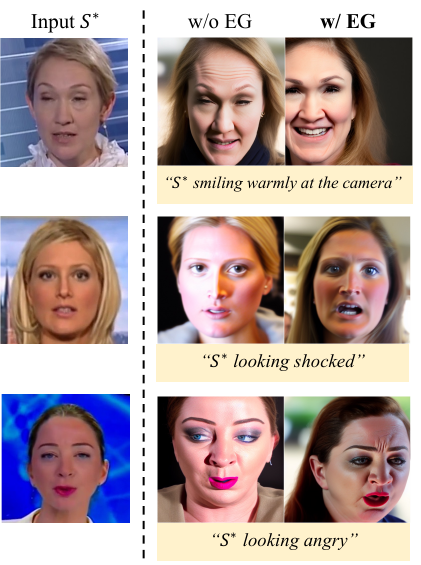}
    \subcaption{Effect of expression guidance.}
    \label{fig:ablation_eg}
  \end{minipage}
  \begin{minipage}{.43\textwidth}
    \centering
    \includegraphics[height=5.75cm]{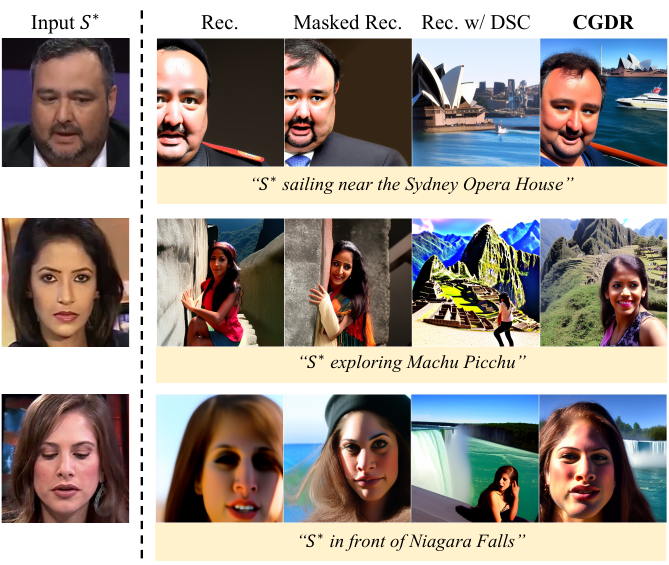}
    \subcaption{Effect of CGDR.}
    \label{fig:ablation_loss}
  \end{minipage}
  \caption{\textbf{Comprehensive ablation study.} We omit the common prefix\textit{``A photo of''} due to the space limitation. (a) Our MSID encoder disentangles camera poses while keeping the identity similarity. (b) Our expression guidance mitigates overfitting to input face expressions. (c) Our CGDR improves the text-fidelity mainly on backgrounds.}
  \label{fig:ablation}
\end{figure*}

\subsection{Comparison with Previous Methods}
\noindent\textbf{Qualitative result.}
We show the generated samples by personalization methods in Fig.~\ref{fig:comparison}.
Some of the subject-driven methods such as DreamBooth~\cite{dreambooth} and CustomDiffusion~\cite{customdiffusion} that require per-face otimization generate highly similar faces to input ones. 
However, they struggle with overfitting to identity-irrelevant attributes of input images, which results in ignoring some text conditions, \eg, \textit{``in a vintage boutique''}.
On the other hand, CelebBasis~\cite{celebbasis} and FastComposer~\cite{fastcomposer}, that are specialized for face personalization, are better-aligned with input texts but comprise the identity-fidelity due to their strong regularizations. 
Also, we observe that FastComposer tends to produce inconsistent images between foregrounds and backgrounds, as reported in a previous study~\cite{celebbasis}. This is because DSC adopted in the method switches the input prompts in the middle of the denoising loop.
Compared to these methods, our Face2Diffusion consistently satisfies both input face identities and text conditions, which demonstrates the effectiveness of our approach.
More results are included in the supplementary material.

\begin{table}[t]
    \centering
    \begin{adjustbox}{width=0.95\linewidth}
    \begin{tabular}{lcc|c} \toprule
      
      Encoder&AdaFace & CLIP&hMean\\
      \midrule
        ArcFace ~\cite{arcface}&0.2421 & \textbf{0.2758} & \underline{0.2135}\\
        ArcFace w/ Multi-Scale Feat.~\cite{dreamidentity} &  \textbf{0.3264} & 0.2069 & 0.1549\\
        \rowcolor[gray]{0.9}
        MSID Encoder (Ours) & \underline{0.3143} & \underline{0.2486} & \textbf{0.2252}\\
      \bottomrule
    \end{tabular}
    \end{adjustbox}
  \caption{\textbf{Effect of MSID encoder.} The original ArcFace~\cite{arcface} remains at a low identity similarity. The multi-scale features~\cite{dreamidentity} trades the text-fidelity for the identity-fidelity. Our method improves the text-fidelity while preserving the identity-fidelity.}
  \label{tb:ablation_encoder}
\end{table}

\noindent\textbf{Quantitative result.}
Then, we evaluate the methods quantitatively in Table~\ref{tb:ffpp}.
CustomDiffusion~\cite{customdiffusion} consistently achieves the best results in the identity metrics (AdaFace, SphereFace, and FaceNet); however, it faces the poor editablity due to overfitting to input images.
In contrast, CelebBasis~\cite{celebbasis} and FastComposer~\cite{fastcomposer} obtain better editability compared to others but generated faces have less similarity to input faces. 
As a result, these methods remain at low Identity$\times$Text scores.
Our method ranks in the top-3 in five of the six metrics and outperforms the state-of-the-art methods in Identity$\times$Text scores. 
This result clearly indicates that our model improves the trade-off between identity- and text-fidelity.

We also report FID~\cite{fid}, the numbers of trained parameters, and encoding time on a single NVIDIA A100 GPU in Table~\ref{tb:param_fid} for more comparisons of the top models in Table~\ref{tb:ffpp}. 
Because there is no target distribution (real images) for the test prompts, we compute FID between the generated images and human-centric images whose captions include a human-related word (\textit{``person''},\textit{``man''}, or \textit{``woman''}) from the CC12M~\cite{cc12m} dataset.
We observe that our method achieves the best FID, demonstrating its higher-fidelity and diversity than the previous state-of-the-arts.
For the computation costs, although CelebBasis has less trainable parameters than the others, it is much slower than our method because CelebBasis requires optimization per subject. Our method achieves the fastest encoding time of the methods.

\subsection{Ablation Study and Analysis}

\begin{table}[t]
    \centering
    \begin{adjustbox}{width=0.95\linewidth}
    \begin{tabular}{lcc|c} \toprule
      
      Method&AdaFace & CLIP&hMean\\
      \midrule
        w/o Expression Guidance &\textbf{0.3338} & 0.2315 & 0.2032\\
        \rowcolor[gray]{0.9}
        w/ Expression Guidance (Ours) & 0.3143 & \textbf{0.2486} & \textbf{0.2252}\\
      \bottomrule
    \end{tabular}
    \end{adjustbox}
  \caption{\textbf{Effect of expression guidance.}}
  \label{tb:ablation_eg}
\end{table}

\noindent\textbf{Effect of MSID encoder.}
First, we compare our MSID encoder with the original ArcFace~\cite{arcface} and its multi-scale features~\cite{dreamidentity} in Fig.~\ref{fig:ablation}(b) and Table~\ref{tb:ablation_encoder}.
We observe that the original ArcFace~\cite{arcface} cannot represent subjects' identities sufficiently (0.2421 in AdaFace). Also, it sometimes mistakes subjects' gender because of the excessively abstracted identity features from the deepest layer, as shown in the third row of the figure.
On the contrary, multi-scale features~\cite{dreamidentity} cannot disentangle camera poses of input images, resulting in a low editability (0.2069 in CLIP).
Our MSID encoder successfully improves the editability (from 0.2069 to 0.2486 in CLIP) by preventing overfitting to camera poses while maintaining the benefits of leveraging multi-scale features (0.3143 vs. 0.3264 in AdaFace) and achieves the best Identity$\times$Text score (0.2252 in hMean).

\begin{table}[t]
    \centering
    \begin{adjustbox}{width=0.9\linewidth}
    \begin{tabular}{lcc|c} \toprule
      
      Method&AdaFace & CLIP&hMean\\
      \midrule
        Reconstruction &\underline{0.3133} & 0.2053 & 0.1411\\ 
        Masked Reconstruction & 0.3012 & 0.1990 & 0.1282\\
        Reconstruction w/ DSC~\cite{fastcomposer}&0.1651 & \textbf{0.2851} & \underline{0.1596}\\
        \rowcolor[gray]{0.9}
        CGDR (Ours) & \textbf{0.3143} & \underline{0.2486} & \textbf{0.2252}\\
      \bottomrule
    \end{tabular}
    \end{adjustbox}
  \caption{\textbf{Effect of CGDR.} 
  The best and the second-best values are in
  \textbf{bold} and \underline{underlined}, respectively.
  The original reconstruction training and simple masking strategy result in the low editability. 
  DSC trades the identity-fidelity for the text-fidelity. Our method balances the trade-off and achieves the best Identity$\times$Text score.}
  \label{tb:ablation_loss}
\end{table}

\noindent\textbf{Effect of expression guidance.}
We evaluate our method with and without expression guidance in Fig.~\ref{fig:ablation}(c) and Table~\ref{tb:ablation_eg}.
We can see that our expression guidance improves the text-fidelity (from 0.2315 to 0.2486 in CLIP) especially on expression-related prompts (\eg, \textit{``$S^*$ looking shocked''}), as shown in the figure. 
We also observe that the identity-fidelity of our model is slightly lower than our model without expression guidance (0.3154 vs. 0.3338 in AdaFace). 
This is because generated images have diverse face expressions aligned with text prompts, which sometimes degrades the quantitative identity similarity of generated images.

\noindent\textbf{Effect of CGDR.}
We compare our CGDR with three baselines in Fig.~\ref{fig:ablation}(a) and  Table~\ref{tb:ablation_loss}:
``Reconstruction (Rec.)'' is our model trained with the original reconstruction loss (Eq.~\ref{eq:loss_rec}) instead of our CGDR (Eq.~\ref{eq:loss_cgdr}). 
``Masked Reconstruction (Masked  Rec.)'' is our model trained with a reconstruction loss computed within facial regions using segmentation masks.
Moreover, we compare Delayed Subject Conditioning (DSC)~\cite{fastcomposer} that uses an alternative prompt whose identifier $S^*$ is replaced with its class name (\ie, \textit{a person}) in early denoising steps during inference. It is partially similar to our CGDR from the perspective of leveraging class-guided denoising. 
We implement this technique on the ``Reconstrucion'' model and denote it as ``Reconstruction w/ DSC  (Rec.~w/ DSC)''.
We observe that the original reconstruction training causes overfitting to training samples, which results in a low editability (0.2053 in CLIP).
And the masked reconstruction training does not solve the overfitting problem (0.1990 in CLIP).
DSC~\cite{fastcomposer} highly improves the text-fidelity (from 0.2053 to 0.2851 in CLIP) but brings a fatal degradation in the identity similarity (from 0.3133 to 0.1651 in AdaFace).
This is because the class prompt \textit{``a person''} is sometimes ignored when emphatic words coexist in the input prompts, as shown in Fig.~\ref{fig:ablation}(a).
Our method improves the text-fidelity from the original reconstruction model (from 0.2053 to 0.2486 in CLIP) and brings a large improvement on the trade-off between identity- and text-fidelity (0.2252 in hMean), which clearly proves the effectiveness of our CGDR.

\begin{figure}[t]
\centering
\includegraphics[width=1.0\linewidth]{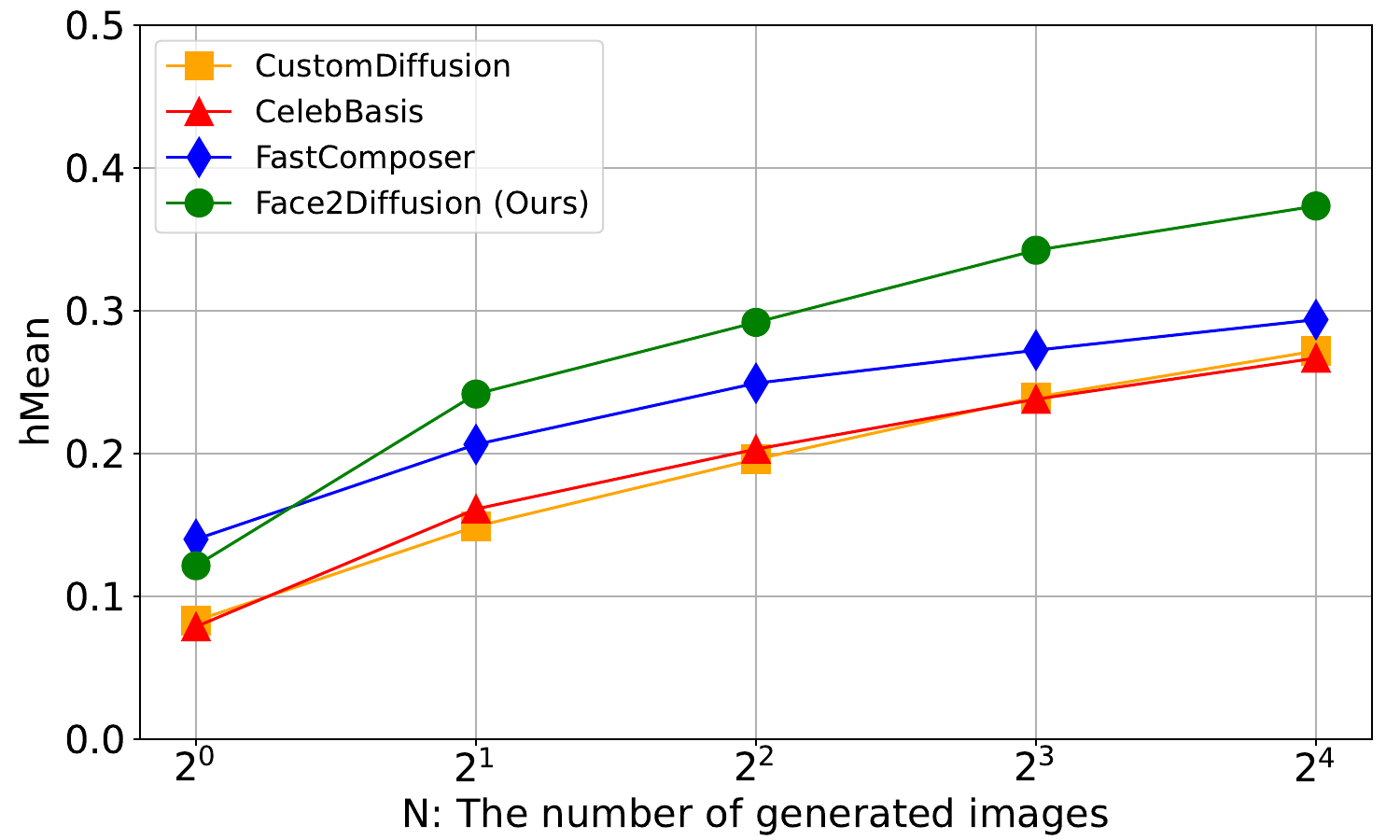} 
\caption{\textbf{Upper-bound analysis.}}
  \label{fig:upperbound}
\end{figure}

\noindent\textbf{Upper-bound analysis.}
Given a practical situation where users can redraw the initial noise several times to produce the most preferable image, it is more important to evaluate only the best image rather than all the generated images.
To meet this demand, we conduct an upper-bound analysis where we evaluate only the single image that achieves the best hMean in the $N$ generated images for each \{image, prompt\} set.
We plot the upper-bound hMean scores when $N=\{1,2,4,8,16\}$ in Fig.~\ref{fig:upperbound}.
We can see that our method surpasses the previous methods by the larger margins as $N$ increases while FastComposer shows a limited improvement in hMean even as $N$ increases.
An important observation from Table~\ref{tb:ffpp} and Fig.~\ref{fig:upperbound} is that FastComposer produces somewhat agreeable results on average, but it is difficult to generate miracle samples. 
On the other hand, our method is more likely to generate very desirable images when the initial noises are redrawn several times.

\section{Limitation}
Because our CGDR leverages the class word (\ie, \textit{``a person''}), our method is sometimes affected by the bias of \textit{``a person''} inherent in T2I models.
For example, we observe that StableDiffusion, which we use in this paper, sometimes produces extremely zoomed out faces when emphatic words (\eg, famous locations) coexist in input prompts; it degrades the identity similarity of generated faces.
For the variants ``Reconstruction'', ``Reconstruction w/ DSC'', and ``CGDR'' in Table~\ref{tb:ablation_loss}, we compute the identity similarity by AdaFace~\cite{adaface} on 10 prompts that include famous locations from our test set and get 0.2963, 0.0663, and 0.2833, respectively. 
Our CGDR is lower than the original reconstruction loss in this case although it outperforms DSC~\cite{fastcomposer}.

\section{Conclusion}
In this paper, we present Face2Diffusion for editable face personalization.
Motivated by the overfitting problem on the previous personalization methods, we propose multi-scale identity encoder, expression guidance, and class-guided denoising regularization to disentangle camera poses, face expressions, and backgrounds from face embeddings, respectively.
We conduct extensive experiments on 100 faces from the FaceForensics++ dataset and 40 text prompts including diverse scenes. The results indicate that our method greatly improves the trade-off between the identity-fidelity and text-fidelity, outperforming previous state-of-the-art methods in the harmonic mean and geometric mean of six metrics that reflect the total quality of face personalization.

\noindent\textbf{Broader Impact.}
Face personalization aims to augment human-centric content creation.
However, our method can be misused for malicious purposes, \eg, to create fake news, as well as previous personalization methods. 
To mitigate the risk, we contribute to the research community by releasing the generated images for image forensics (\eg,~\cite{dire}).


\appendix

\begin{figure}[t]
    \centering
    \includegraphics[width=1.0\linewidth]{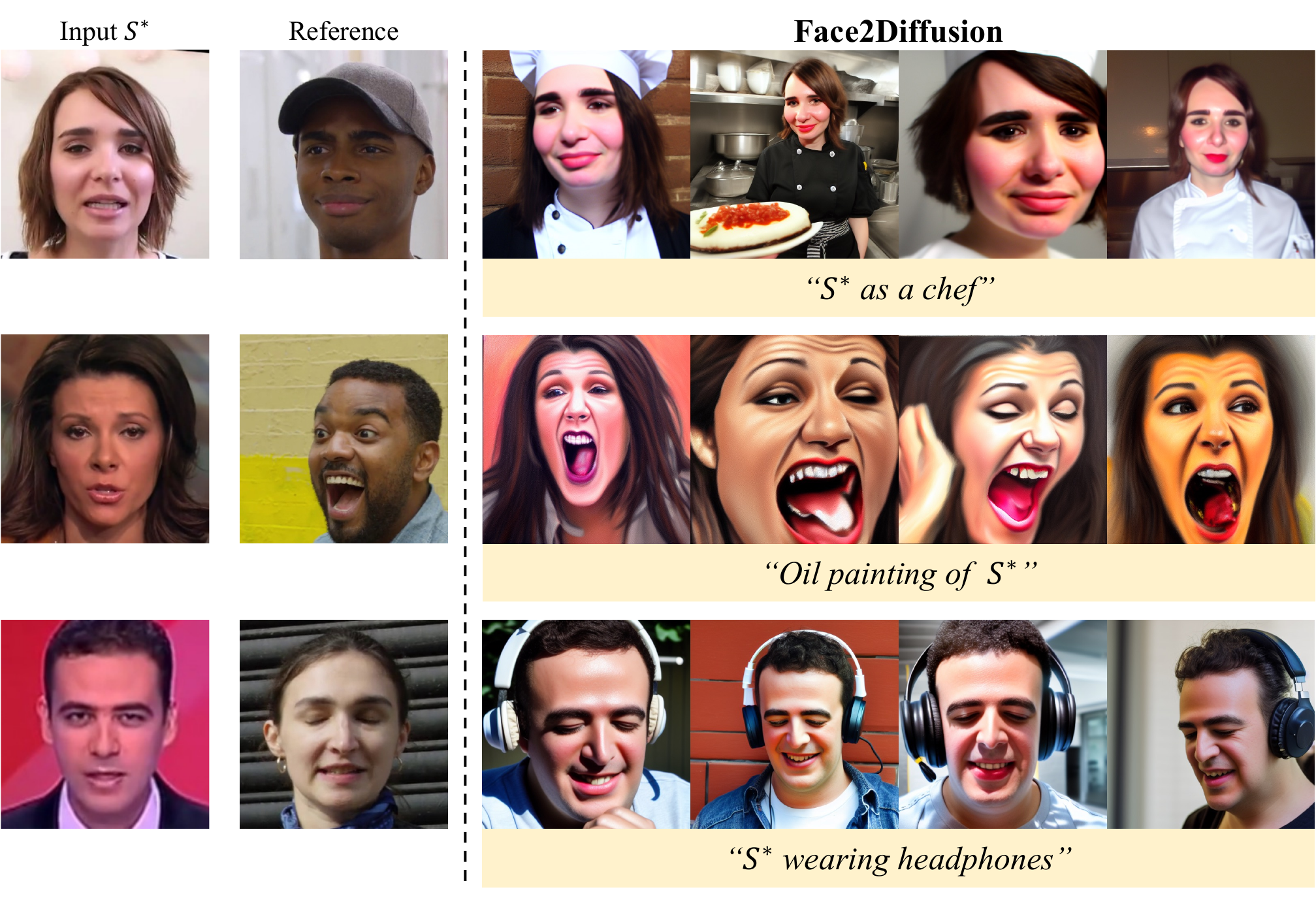}
    \caption{\textbf{Expression-conditional generation.}}
    \label{fig:conditional_additional}
\end{figure}

\section{Expression-Conditional Generation}
Although our expression guidance aims to disentangle face expressions from face embeddings $S^*$, it also enables F2D to generate conditioned face images by reference expressions.
We show the examples of the expression-conditioned generation in Fig.~\ref{fig:conditional_additional}.
The reference images are sampled from the DFD~\cite{dfd} dataset.

\section{Comparison with More Recent Models}
We additionally compare our model with the two recent models, OFT~\cite{oft} and DVAR~\cite{dvar}, in Table~\ref{tb:recent}. 
Our method significantly outperforms such the recent models on hMean and gMean.

\section{Test Prompts}
We give the set of text prompts used in our experiments in Table~\ref{tb:prompt}. 
Our prompts include various scenes related to job, activity, expression, and location.

\section{More Visual Comparisons}
\begin{figure*}[t]
    \centering
    \includegraphics[width=0.8\linewidth]{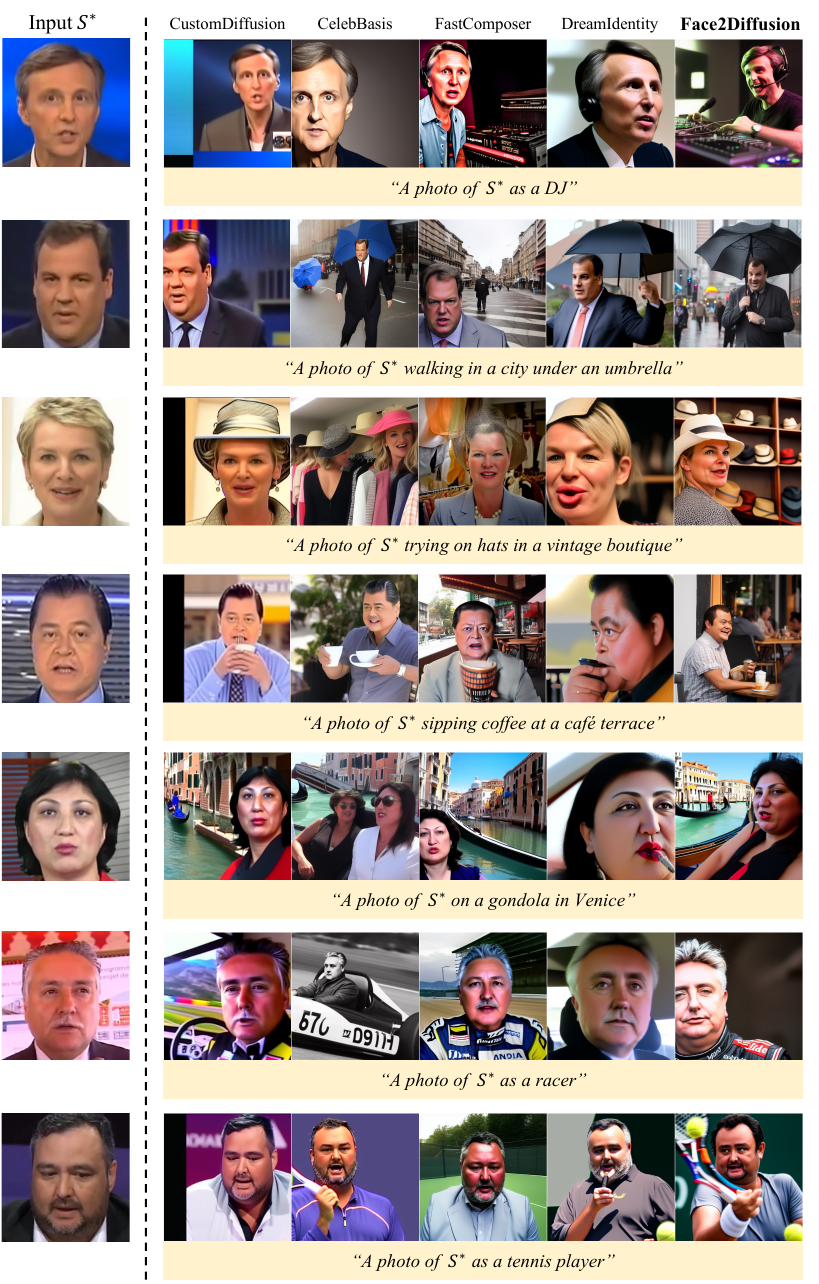}
    \caption{\textbf{Visual comparisons with previous methods.}}
    \label{fig:top5_1}
\end{figure*}

\begin{figure*}[t]
    \centering
    \includegraphics[width=0.8\linewidth]{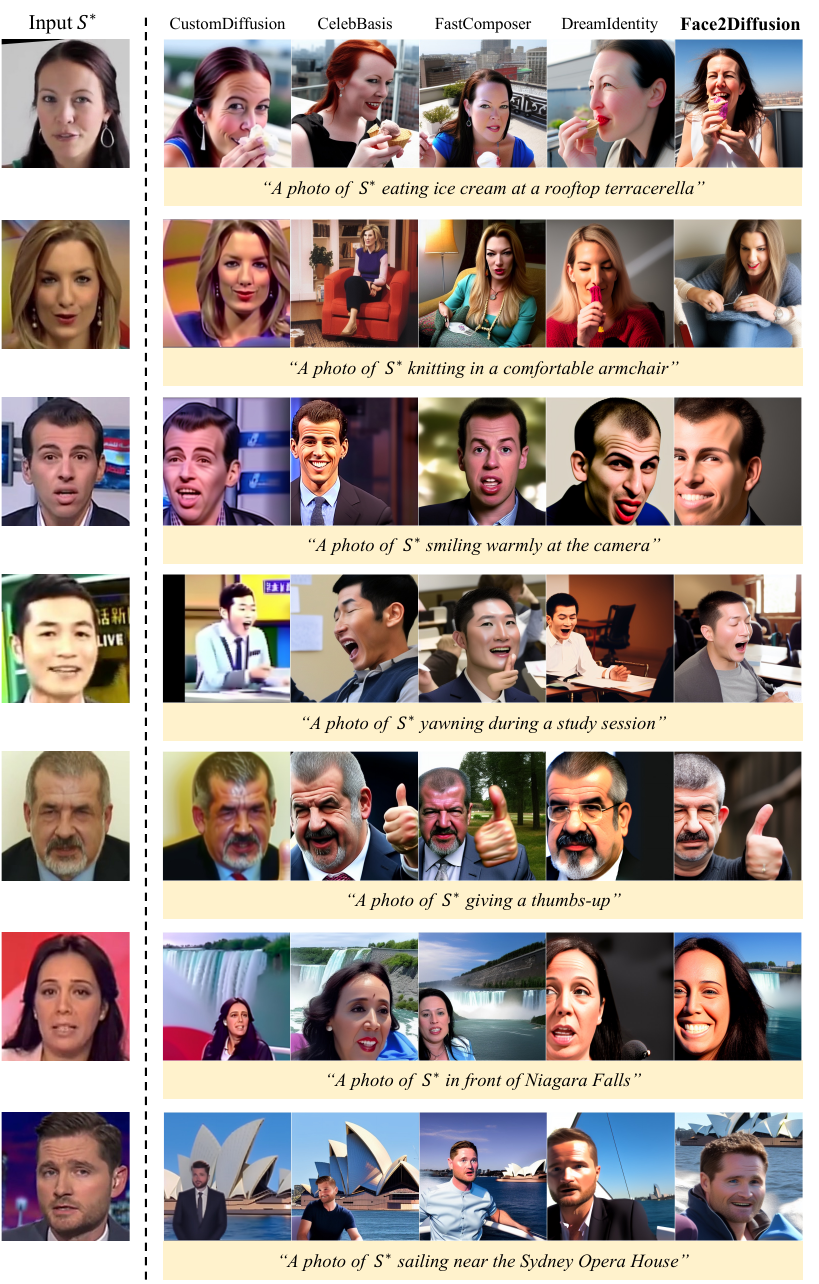}
    \caption{\textbf{Visual comparisons with previous methods.}}
    \label{fig:top5_2}
\end{figure*}

We show additional examples in Figs.~\ref{fig:top5_1} and \ref{fig:top5_2}.
For enhanced visibility, we compare our method with CustomDiffusion~\cite{customdiffusion}, CelebBasis~\cite{celebbasis}, and FastComposer~\cite{fastcomposer} that are ranked in the top-5 in Identity$\times$Text scores in Table~{\color{red}1}.

\begin{table}[t]
    \centering
    \begin{adjustbox}{width=1.0\linewidth}
    \begin{tabular}{lcccccc|cc} \toprule
        &AdaFace& SphereFace& FaceNet&CLIP&dCLIP & SigLIP&hMean&gMean\\ 
      \midrule
      OFT &0.3446 & 0.3980 & 0.4673 & 0.2245 & 0.1364 & 0.3515 & 0.0615 & 0.0993\\
      DVAR &0.0452 & 0.0939 & 0.1201 & 0.2710 & 0.1852 & 0.4261 & 0.0369 & 0.0548\\
      Ours & 0.3143 & 0.4215 & 0.5313 & 0.2486 & 0.2020 & 0.3856 & \textbf{\fontsize{11pt}{12pt}\selectfont 0.1749} & \textbf{\fontsize{11pt}{12pt}\selectfont 0.2252}\\
      \bottomrule
    \end{tabular}
    \end{adjustbox}
  \caption{\textbf{Comparison with the more recent methods.}}
  \label{tb:recent}
\end{table}

\begin{table}[t]
    \centering
    \begin{adjustbox}{width=1.0\linewidth}
    \begin{tabular}{c} \toprule
       Prompts \\ 
      \midrule
        A photo of $S^*$ as a firefighter \\
        A photo of $S^*$ as a cowboy \\
        A photo of $S^*$ as a chef \\
        A photo of $S^*$ as a racer \\
        A photo of $S^*$ as a king \\
        A photo of $S^*$ as a scientist \\
        A photo of $S^*$ as a tennis player \\
        A photo of $S^*$ as a DJ \\
        A photo of $S^*$ as a knight \\
        A photo of $S^*$ as a pilot \\
        A photo of $S^*$ walking in a city under an umbrella \\
        A photo of $S^*$ surrounded by tall bookshelves \\
        A photo of $S^*$ trying on hats in a vintage boutique \\
        A photo of $S^*$ sipping coffee at a café terrace \\
        A photo of $S^*$ in a busy subway station \\
        A photo of $S^*$ eating ice cream at a rooftop terrace \\
        A photo of $S^*$ playing the saxophone on a stage \\
        A photo of $S^*$ running in a meadow \\
        A photo of $S^*$ playing chess at a wooden table \\
        A photo of $S^*$ knitting in a comfortable armchair \\
        A photo of $S^*$ yawning during a study session \\
        A photo of $S^*$ smiling warmly at the camera \\
        A photo of $S^*$ with a hand covering their mouth \\
        A photo of $S^*$ hugging a friend tightly \\
        A photo of $S^*$ flexing muscles in a gym \\
        A photo of $S^*$ looking shocked \\
        A photo of $S^*$ giving a thumbs-up \\
        A photo of $S^*$ looking angry \\
        A photo of $S^*$ sitting cross-legged on a rock \\
        A photo of $S^*$ wearing an oversized sweater \\
        A photo of $S^*$ at the Great Wall of China \\
        A photo of $S^*$ exploring Machu Picchu \\
        A photo of $S^*$ sailing near the Sydney Opera House \\
        A photo of $S^*$ walking through the streets of Rome near the Colosseum \\
        A photo of $S^*$ at the Grand Canyon in sunset \\
        A photo of $S^*$ enjoying cherry blossoms in Tokyo \\
        A photo of $S^*$ on a gondola in Venice \\
        A photo of $S^*$ at the Taj Mahal in India \\
        A photo of $S^*$ at Mount Everest Base Camp \\
        A photo of $S^*$ in front of Niagara Falls \\
      \bottomrule
    \end{tabular}
    \end{adjustbox}
  \caption{\textbf{Our prompt set.}}
  \label{tb:prompt}
\end{table}

\clearpage

\clearpage

\section{Implementation Details}
\subsection{Previous Methods}
To conduct fair comparisons, we implement previous methods by strictly following the official instructions as much as possible. Commonly in existing methods and our Face2Diffusion, we use StableDiffusion-v1.4 (SD1.4)~\cite{latentdiffusion}, Euler ancestral discrete scheduler~\cite{kdiffusion} with 30 denoising steps, and classifier-free guidance~\cite{cfdg} with a scale parameter of 7.0.
Specific details of each method are as follows:

\noindent\textbf{TextualInversion.}
We use the diffusers' implementation~\cite{diffusers}.
The inverted embedding is initialized by \textit{``person''}.

\noindent\textbf{DreamBooth.}
We use the diffusers' implementation~\cite{diffusers}.
For the prior preservation loss, we use other face images from our test set, \ie, 99 identities. We set the class word for the regularization to \textit{``a person''}.

\noindent\textbf{CustomDiffusion.}
We use the official implementation integrated into diffusers~\cite{diffusers}.
We use the same regularization as DreamBooth.

\noindent\textbf{Perfusion.}
We directly use the official training code~\cite{perfusion}.
The inverted embedding is initialized by \textit{``person''}.

\noindent\textbf{E4T.}
We directly use the official training code~\cite{e4t}.

\noindent\textbf{CelebBasis.}
We directly use the official training code~\cite{celebbasis}.

\noindent\textbf{FastComposer.}
We use the official implementation~\cite{fastcomposer}.
Because the released checkpoint is based on SD1.5, we train it from scratch on SD1.4 using the official training code.
We use \textit{''a person''} for delayed subject conditioning (DSC).

\noindent\textbf{ELITE.}
We directly use the official pretrained model~\cite{elite}.
For segmentation masks during inference, we use a face-parsing model~\cite{faceparsing} that is the same one used in our CGDR.

\noindent\textbf{DreamIdentity.}
Because there is no public implementation, we re-implement it.
For the mapping MLP, we use the same architecture as our Face2Diffusion because the implementation detail is not described in the original paper.
Due to the limitations of our computational resource, we train the model with eight NVIDIA A100 (40GB) GPUs which is the same cost as our F2D though the original paper~\cite{dreamidentity} use its 80GB version.
For self-augmented data, we collect 1K celebrity names from Internet that are consistently generated by SD1.4. Because some of proposed editing prompts do not work on SD1.4, we remove them and add alternative ones tested in the original paper. 
In total, we generate 8K augmented images (1K identities $\times$ 8 editing prompts).

\subsection{Variants of F2D}
\noindent\textbf{Reconstruction.}
We use the same loss as Eq.~{\color{red}1}.

\noindent\textbf{Masked Reconstruction.}
We use a masked reconstruction loss as follows:
\begin{equation}
\label{eq:loss_mask}
 \mathcal{L}=\lVert (\epsilon -  \epsilon_{\theta}(z_t,t,\tau(p))) \odot M \rVert_2^2.
\end{equation}

\noindent\textbf{Reconstruction w/ DSC.}
We implement DSC~\cite{fastcomposer} on the ``Reconstruction'' model above. 
Following the official implementation, we adopt the  ratio of $\alpha=0.8$ for DSC. 

\noindent\textbf{ArcFace.}
We implement ViT~\cite{vit} trained with ArcFace loss~\cite{arcface} using an unofficial implementation~\cite{insightface}. We input only the deepest layer's outputs corresponding the classifier token into the mapping network $f_{map}$.

\noindent\textbf{ArcFace w/ MSF.}
We extract multi-scale features (MSF)~\cite{dreamidentity} from the ArcFace model above.
We use the same depth set as our F2D for MSF, \ie, $\{3,6,9,12\}$.

\noindent\textbf{w/o Expression Guidance.}
We remove the concatenation before the mapping network. Therefore, the identifier $S^*$ is computed during both training and inference as follows:
\begin{equation}
  S^* = f_{map}(f_{id}(x)).
\end{equation}

\noindent\textbf{ControlNet.}
We adopt an unofficial implementation~\cite{controlnet_repo} of ControlNet for facial landmarks.
Because the pretrained model is built on SD1.5, we train our model without expression guidance on SD1.5 and then we combine them.

\subsection{Metrics}
\noindent\textbf{AdaFace/SphereFace/FaceNet.}
We use the official and unoffical implementations~\cite{adaface,sphereface_repo,facenet_repo}.
We compute the cosine similarity between extracted features of an input image $x$ and generated image $y$, and then clip the value to $[0, 1]$:
\begin{equation}
\text{ID} = \max(\cos(f_{fr}(x),f_{fr}(y)),0),
\end{equation}
where $\cos$ and $f_{fr}$ represent the cosine similarity and each feature extractor of face recognition models, respectively.

\noindent\textbf{CLIP.}
The CLIP score~\cite{clipscore} evaluates the cosine similarity between a generated image and input prompt $p$:
\begin{equation}
\text{CLIP} = \max(\cos(E_{I}(y), E_{T}(p)),0),
\end{equation}
where $E_{I}$ and $E_{T}$ are CLIP image and text encoders, respectively.
We use the official implementation~\cite{clip} of CLIP ViT-H/14 model trained on the LAION-2B~\cite{laion} dataset.

\noindent\textbf{dCLIP.}
The directional CLIP (dCLIP) score~\cite{nada} evaluates the cosine similarity between the difference vectors from the reference points in the image and text space.
We set the reference prompt $p_r$ to ``\textit{A photo of a person}'' and the reference image $y_r$ to an image generated by ``\textit{A photo of $S^*$}''. The dCLIP score is computed as:
\begin{eqnarray}
&\text{dCLIP} = \max(\cos(\Delta{y}, \Delta{p}),0),\\
&\Delta{x} = E_{I}(y) - E_{I}(y_r), \quad \Delta{p} = E_{T}(p) - E_{T}(p_r).
\end{eqnarray}
We use the same encoders as the CLIP score.

\noindent\textbf{SigLIP.}
The SigLIP score is a scaled cosine similarity between an image and text, which is computed as:
\begin{equation}
\text{SigLIP} = \sigma(s\cdot \cos(E_{I}(y), E_{T}(p)) + b),
\end{equation}
where $s$ and $b$ are the scale and bias parameters optimized during the pretraining of SigLIP.
$\sigma$ represents the sigmoid function.
We use the official implementation~\cite{siglip} of SigLIP trained on the WebLI~\cite{webli} dataset.

\section*{Acknowledgements}
This work is supported by JSPS KAKENHI Grant Number JP23KJ0599 and partially supported by JSPS KAKENHI Grant Number JP22H03640 and Institute for AI and Beyond of The University of Tokyo.

{
    \small
    \bibliographystyle{ieeenat_fullname}
    \bibliography{main}
}


\end{document}